\documentclass[conference]{IEEEtran}
\IEEEoverridecommandlockouts
\usepackage{cite}
\usepackage{amsmath,amssymb,amsfonts}
\usepackage{algorithm}
\usepackage{algpseudocode}
\usepackage{graphicx}
\usepackage{textcomp}
\usepackage{xcolor}
\usepackage{array}
\usepackage{booktabs}
\usepackage{multirow}

\def\BibTeX{{\rm B\kern-.05em{\sc i\kern-.025em b}\kern-.08em
    T\kern-.1667em\lower.7ex\hbox{E}\kern-.125emX}}

\usepackage{xcolor}
\usepackage{soul}

\newcounter{BONumberOfComments}
\stepcounter{BONumberOfComments}

\usepackage{tikz}
\newcommand{\designlabel}[1]{%
    \tikz[baseline=(char.base)] \node[shape=circle, fill=black, draw, inner sep=1pt] (char) {\textcolor{white}{\footnotesize{#1}}};%
}

\begin{document}

\title{EcoLens: Leveraging Multi-Objective Bayesian Optimization for Energy-Efficient Video Processing on Edge Devices\\
\thanks{This work was in part supported by NSF CCF 22-17144, NSF IIS 21-40645, NSF 21-06592, NSF CNS 19-00875, Fisher Professorship, and Sony gift funds.}
}

\author{
    \IEEEauthorblockN{Benjamin Civjan, Bo Chen, Ruixiao Zhang, Klara Nahrstedt}
    \IEEEauthorblockA{\textit{Siebel School of Computing and Data Science}\\
    \textit{University of Illinois Urbana-Champaign}\\
    Urbana, IL, USA\\
    Email: \{bcivjan2, boc2, ruixiao, klara\}@illinois.edu}
}

\maketitle

\begin{abstract}
Video processing for real-time analytics in resource-constrained environments presents a significant challenge in balancing energy consumption and video semantics. This paper addresses the problem of energy-efficient video processing by proposing a system that dynamically optimizes processing configurations to minimize energy usage on the edge, while preserving essential video features for deep learning inference. We first gather an extensive offline profile of various configurations consisting of device CPU frequencies, frame filtering features, difference thresholds, and video bitrates, to establish apriori knowledge of their impact on energy consumption and inference accuracy. Leveraging this insight, we introduce an online system that employs multi-objective Bayesian optimization to intelligently explore and adapt configurations in real time. Our approach continuously refines processing settings to meet a target inference accuracy with minimal edge device energy expenditure. Experimental results demonstrate the system’s effectiveness in reducing video processing energy use while maintaining high analytical performance, offering a practical solution for smart devices and edge computing applications.
\end{abstract}


\section{Introduction}
\begin{figure}[t]
    \centering
    \includegraphics[width=0.5\textwidth]{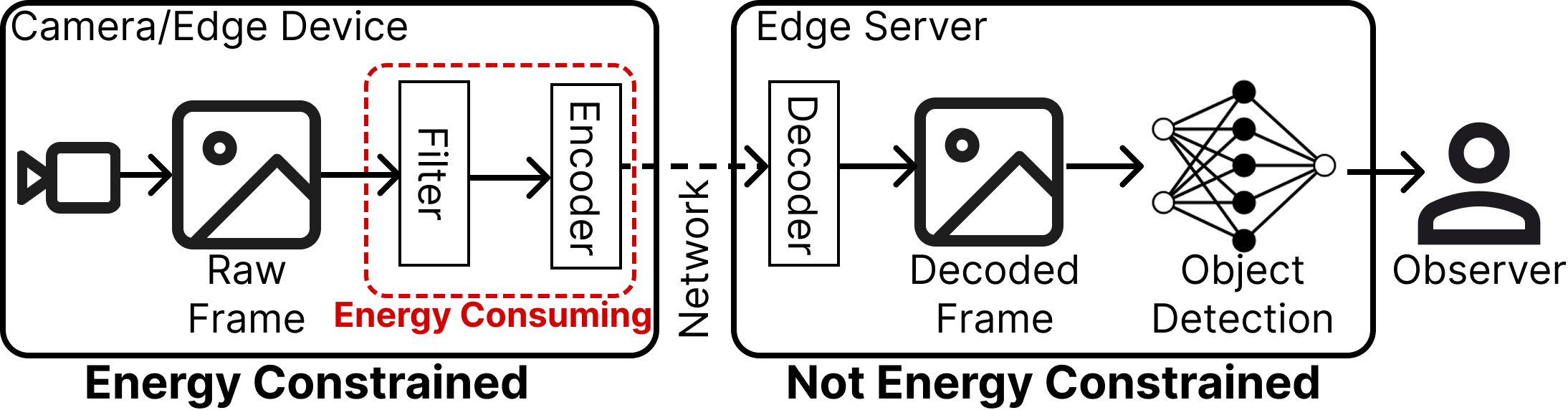}
    \caption{Overview of a generic video analytics pipeline. In this paper we focus on the energy consumption of the edge device video processing, shown in red.}
    \label{fig:analytics-pipeline}
\end{figure}

Video camera deployments are more prevalent than ever and capture massive amounts of video data in a wide range of settings, including city surveillance, wearable technology, and first-responder monitoring~\cite{chicago_surveillance, jh_livecam, alma_livecam, ring_security, meta_ai_glasses, firefighter_training, st360}. Due to the vast amount of live information captured from these cameras, automated pipelines increasingly rely on deep neural networks (DNNs) running on edge servers for fast and accurate content analysis (Figure~\ref{fig:analytics-pipeline}). Extensive research has been conducted to optimize the inference accuracy, speed, and bandwidth efficiency of these pipelines~\cite{serverinference, reducto, recl, edgeanalyticssurvey}.

A common approach is to use \textit{on-camera} filtering to eliminate redundant frames before transmission from camera to server~\cite{reducto, adaframe, glimpse}. For example, one state-of-the-art work~\cite{reducto} evaluates light-weight image features (\texttt{pixel}, \texttt{area}, and \texttt{edge}) on the camera itself and chooses whether to send the frame based on a difference threshold (ranging from 0 - 100\%). This technique offers two primary benefits: reducing bandwidth usage and decreasing computational demand on the edge server. However, the focus of this on-camera filtering has traditionally been on bandwidth and server computation reduction, which ignores edge device energy consumption.

Many streaming pipelines involve energy-constrained edge devices such as battery-powered cameras\cite{smartcam_energy, firefighter_training, ring_security}, smart glasses~\cite{meta_ai_glasses}, and drones~\cite{drone_streaming}, where optimizing energy efficiency is critical for prolonged operation. For instance, battery-powered cameras used in search-and-rescue firefighter missions perform continuous video streaming for situational awareness while also needing to conserve battery life to maximize operation time. Similarly, smart glasses, drones, and home security cameras are deployed for real-time tasks like intruder detection, hazard monitoring, and surveillance, yet must operate within strict energy limits. In these applications, inefficient video processing can lead to frequent recharging or device downtime, reducing system usability and effectiveness. We discuss more about specific device lifetimes in~\ref{limited-device-energy}.

Designing an energy-efficient video processing system for edge devices presents a few key challenges. First, there is a complex relationship between the settings used for frame filtering and encoding (Figure~\ref{fig:analytics-pipeline}), DNN inference accuracy, and edge device energy consumption. While reducing frame quality can conserve energy, it may reduce accuracy due to degraded image quality. Similarly, choosing a stricter filter threshold can reduce encoder processing load but may discard useful frames, impacting downstream analysis. Balancing these trade-offs dynamically is non-trivial. Second, the precise relationship of these control knobs changes with the video scene. Therefore, performing a purely offline grid search to find the best configuration is impractical. Selecting the optimal configuration online is also difficult because of the real-time requirements of live streaming, the significant time it takes to profile a configuration, and the volatile nature of video semantics.

To address these challenges we propose \textit{EcoLens}, an energy-aware on-camera filtering system that dynamically configures video processing settings to minimize energy consumption while meeting a target inference accuracy threshold. Our methodology consists of two key phases: offline energy sensing and online configuration selection. First, we gather a device profile that maps each processing configuration to its corresponding energy consumption and inference accuracy for bounding-box queries on an edge server. Leveraging this data, we model the configuration search as a multi-objective Bayesian optimization (MBO) problem and design an online algorithm to dynamically select the optimal video processing configuration to minimize energy usage while meeting the desired accuracy target.

Our contributions are as follows:

\begin{enumerate}
    \item We identify that existing approaches compromise energy efficiency in video streaming pipelines and evaluate the impact of video processing configurations on battery life, finding the best CPU frequency and video filtering feature to use to prioritize edge device energy consumption.
    \item We design and propose an energy-efficient on-camera filtering system, EcoLens, that leverages offline sensing and MBO for online selection to reduce energy consumption while preserving inference accuracy.
    \item We conduct an extensive evaluation on multiple real-world live video streams, comparing the performance of EcoLens to a variety of different video processing methods. Our evaluation of EcoLens demonstrates edge device energy savings of up to 44.6\% compared to other methods. Additionally, EcoLens continually maintains downstream analytics accuracy, staying within 3\% of the target value. Source code and evaluation data for EcoLens are available at https://github.com/bencivjan/ecolens.
\end{enumerate}

\section{Related Work and Background}

\subsection{On-Camera Content Filtering}
\label{filtering-background}
On-camera content filtering aims to reduce the amount of data transmitted from edge devices to remote servers by selectively processing frames before transmission. AdaFrame~\cite{adaframe} adaptively selects frames for video recognition using an LSTM-based policy network with a global memory, determining which frames to process based on predicted utility. It dynamically adjusts frame selection per video. Glimpse~\cite{glimpse} takes a different approach to frame filtering. It takes the pixel difference between frames and uses a static difference threshold to determine which frames to send to the server. Reducto~\cite{reducto} expands on Glimpse, evaluating multiple low-level frame features (e.g., pixel, area, and edge) to determine the best feature for different query types. Similar to Glimpse, it filters frames directly on the camera and transmits only frames that exceed a predefined threshold. However, Reducto dynamically tunes the differencing threshold to continuously meet the analytics accuracy requirement for a range of different videos.

While these techniques focus on reducing bandwidth and server computation costs, they neglect edge device energy. Our work, EcoLens, builds on these approaches by introducing an energy-aware filtering mechanism that dynamically adjusts video processing configurations to minimize edge power consumption while maintaining target inference accuracy.

\subsection{Multi-objective Bayesian Optimization}
\label{MBO}
Multi-objective Bayesian optimization (MBO) is a powerful technique for optimizing multiple conflicting objectives in expensive black-box functions, making it widely applicable in machine learning, engineering design, and resource-constrained systems \cite{bofl, onlinembo, mboglobal}. In the context of video processing, MBO can be used to efficiently select configurations that balance competing trade-offs such as inference accuracy and energy consumption. MBO selects points in a way that expands the \textbf{Pareto front} of the objectives. A solution is Pareto-optimal if no other solution improves one objective without worsening at least one other objective.

\section{Motivation and Challenges}

\begin{figure}[t]
    \centering
    \includegraphics[width=0.5\textwidth]{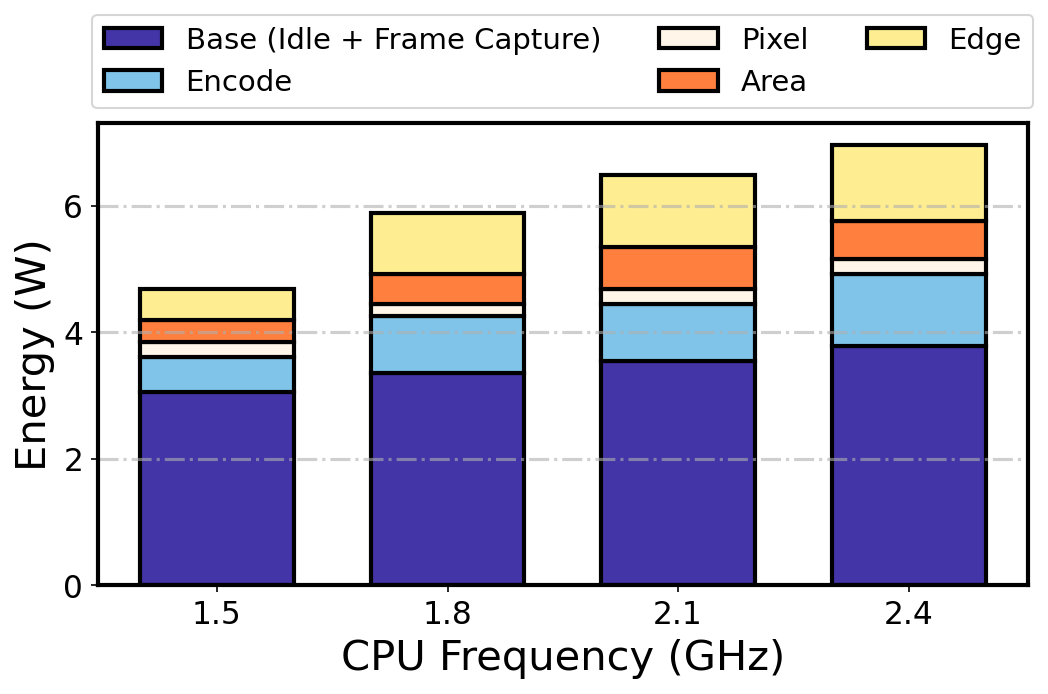}
    \caption{Energy usage by each stage of the video encoding pipeline shown as a function of the CPU frequency. The experiment was performed at 15 fps using a single threaded implementation for consistency.}
    \label{fig:motivation-1-single}
\end{figure}

\begin{table}[t]
    \caption{Device Battery Life Estimated from Raspberry Pi Energy Usage}
    \centering
    \renewcommand{\arraystretch}{1.2} 
    \begin{tabular}{l c c}
        \toprule
        \textbf{Device} & \textbf{Battery Capacity (mAh)} & \textbf{Video Battery Life} \\ \midrule
        Meta Ray-Ban & 154 & 6.6 minutes \\
        GoPro Hero13 & 1,900 & 81 minutes \\
        Mid-range Drone & 5,000 & 3.6 hours \\
        Portable Battery & 10,000 & 7.1 hours \\
        \bottomrule
    \end{tabular}
    \label{tab:battery_life}
\end{table}

\subsection{Limited Edge Device Energy}
\label{limited-device-energy}


As mentioned previously, many video analytics pipelines are limited by the energy requirements of the camera. Table \ref{tab:battery_life} presents the battery capacity of four distinct video recording devices. We calculate the video battery life based on the device battery life using an estimated video processing energy consumption of 7 W, a conservative estimate based on our experiments with a Raspberry Pi 5~\cite{rpi5}. This is not a precise measurement of the device's operational lifetime; rather, it is an estimation of the device's battery lifetime while being used to power a representative edge device. It is clear that each device has a limited energy budget, and the video processing configuration (filtering and encoding) it uses plays a key role in the overall operational time. Prolonging the battery life of these devices while processing video allows for longer capture of content and saves costs.

\subsection{Deciding Control Knobs}

\begin{figure}[t]
    \centering
    \includegraphics[width=0.5\textwidth]{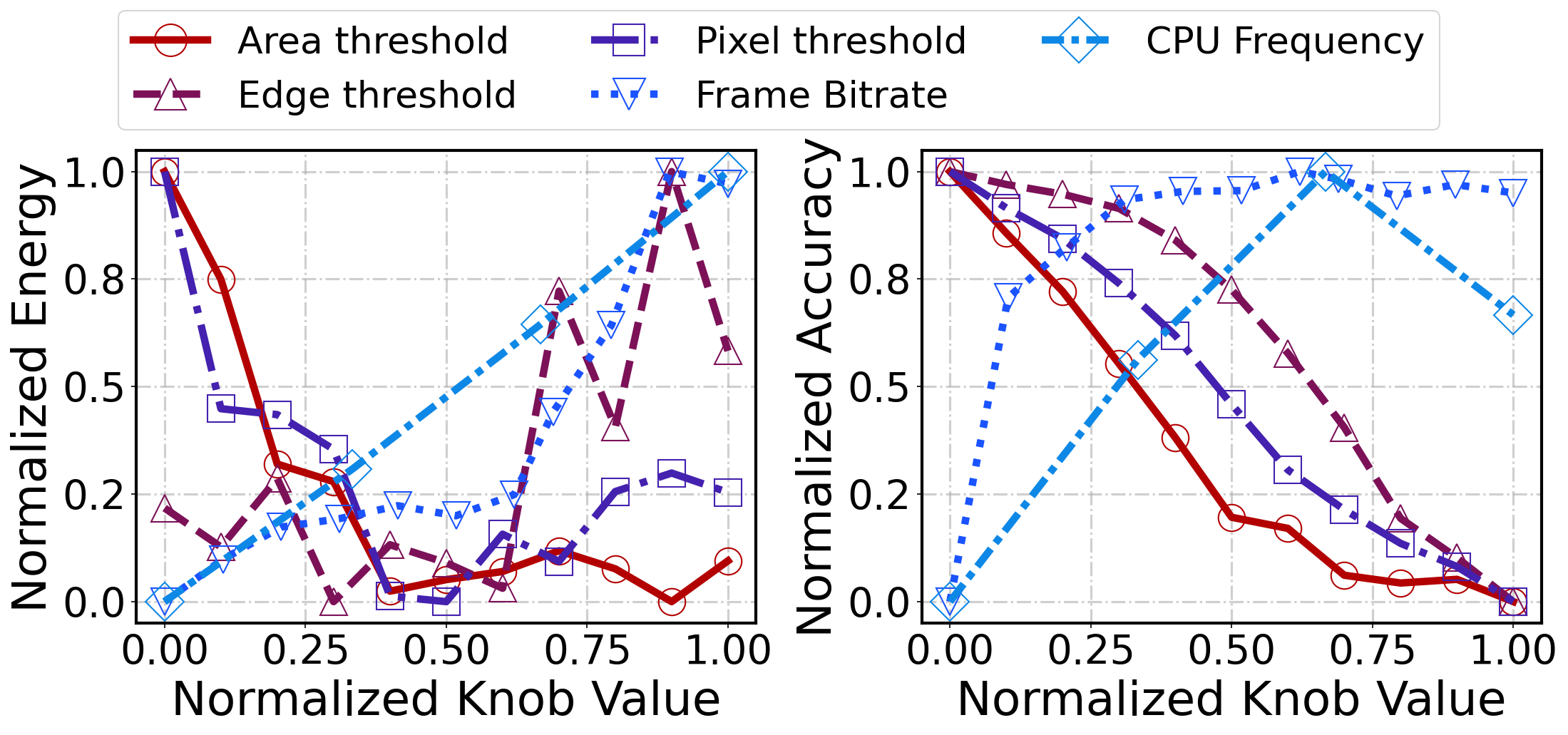}
    \caption{The impact of each control knob on the accuracy and energy is very complex. Generally, as the energy consumption of a configuration decreases, the accuracy also decreases.}
    \label{fig:feature-complexity}
\end{figure}

\begin{figure}[t]
    \centering
    \includegraphics[width=0.5\textwidth]{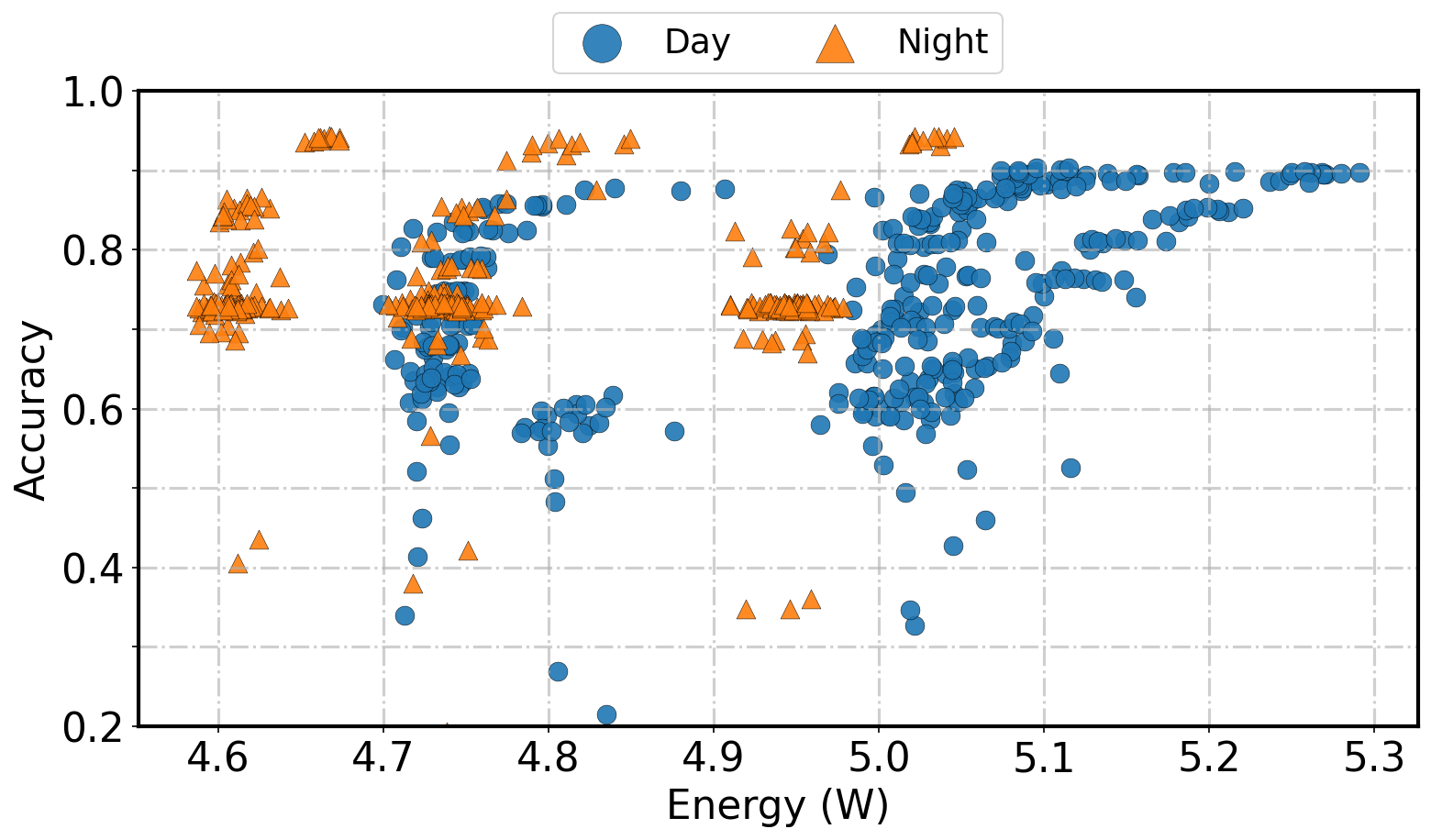}
    \caption{A subset of the sensing results on the Jackson Hole live camera~\cite{jh_livecam}. As the scene changes from day to night, the distribution of configurations in objective space drifts.}
    \label{fig:motivation-day-vs-night}
\end{figure}

In order to optimize the energy/accuracy trade-off of video processing on the edge, we first need to find the parameters of video processing that reduce energy (i.e. our control knobs). Figure~\ref{fig:motivation-1-single} shows the energy consumed during the encoding and filtering stages when CPU frequency is increased. We examine the same feature filters explored in Reducto~\cite{reducto}; \texttt{pixel}, the pixel difference between frames, \texttt{area}, the area difference between frames, and \texttt{edge}, the difference in contours of objects. Both the CPU frequency and filter type play a role in the amount of energy usage. Additionally, Figure~\ref{fig:feature-complexity} shows an increasing energy cost when encoding a frame with higher bitrate. We can also see a general trend that the filter difference threshold impacts both energy and inference accuracy, because a lower threshold means more frames are encoded and transmitted to the server. It is worth noting that sending the frames over Wi-Fi consumed a negligible amount of energy compared to filtering and encoding, which is why we do not focus on the cost of networking.

Our control knob values for the edge device are shown in Table~\ref{tab:configurations}. We decide these values carefully. CPU dynamic voltage and frequency scaling (DVFS) on a Raspberry Pi 5\cite{rpi5} allows frequencies in the range of 1.5--2.4 GHz. The real-time filters are based on state-of-the-art video analytics work~\cite{reducto}. Finally, the filter threshold and bitrate ranges were found through experimentation; increasing the filter threshold range would allow almost no frames and increasing the bitrate range would have little perceptible impact on frame quality.

While these specific values were chosen for the Raspberry Pi 5, they can be adjusted to fit other devices and scenarios. For example, the bitrate and filter threshold ranges can be expanded to benefit the system with a larger search space, at the cost of a longer offline sensing time.


\subsection{Feature Complexity and Data Drift Challenges}
\label{profiling-challenges}

\begin{table}[t]
\setlength{\arrayrulewidth}{0.5mm}
\setlength{\tabcolsep}{20pt}
    \caption{Video Profiling Features}
    \centering
    \renewcommand{\arraystretch}{1.25} 
    \begin{tabular}{m{3cm} p{3cm}}
        \toprule[1pt]
        \textbf{CPU Frequency (GHz)} & 1.5, 1.8, 2.1, 2.4 \\ \midrule[1pt]
        \textbf{Filter} & Pixel, Area, Edge \\ \midrule[1pt]
        \textbf{Filtering Threshold} & 0.00, 0.01, 0.02, 0.03, 0.04, 0.05, 0.06, 0.07, 0.08, 0.09, 0.10 \\ \midrule[1pt]
        \textbf{Frame Bitrate (Kbps)} & 100, 400, 700, 1000, 1300, 1600, 1900, 2100, 2400, 2700, 3000 \\ \bottomrule[1pt]
    \end{tabular}
    \label{tab:configurations}
\end{table}


Tuning the control knobs for video streaming to balance energy efficiency and accuracy is a complex challenge. The design space is highly non-linear, meaning that small changes in settings like filter threshold, frame bitrate, or CPU frequency can lead to big and sometimes unpredictable shifts in both energy consumption and inference accuracy. As shown in Figure~\ref{fig:feature-complexity}, different knob settings don’t always follow a clear pattern. For example, increasing frame bitrate might boost accuracy, but it also increases power usage. Similarly, applying aggressive filtering could save energy but reduce overall accuracy due to missing frames. To make things even trickier, these control knobs don’t act in isolation. Certain settings might work well together, while others could cancel out potential benefits. For instance, a higher CPU frequency might help recover some accuracy lost from aggressive filtering, but it also increases power consumption.

A second challenge in choosing the ideal configuration is the dependence on video scene. This means that for each camera setting, it is necessary to reexamine each individual configuration to determine its corresponding energy usage and accuracy. To make matters more difficult, the performance of a configuration can change \textit{within the same camera setting} over time. Figure~\ref{fig:motivation-day-vs-night} shows this drift; as the scene changes from day to night, the energy/accuracy distribution of individual configurations drastically changes. For example, the [1.5 GHz, \texttt{pixel} filter, 0.01 filter threshold, 100 Kbps bitrate] configuration has a downstream inference accuracy of 86.06\% during the day compared to 43.61\% during the night, and consumed an average of 4.97 W during the day compared to 4.62 W at night. The [2.4 GHz, \texttt{area} filter, 0.02 filter threshold, 1300 Kbps bitrate] configuration, on the other hand, has an inference accuracy of 83.14\% during the day compared to 74.19\% at night, yet consumes an average of 7.24 W compared to 7.37 W at night.



\section{System Design} \label{design}


\subsection{EcoLens System Overview}
\begin{figure}[t]
    \centering
    \includegraphics[width=0.5\textwidth]{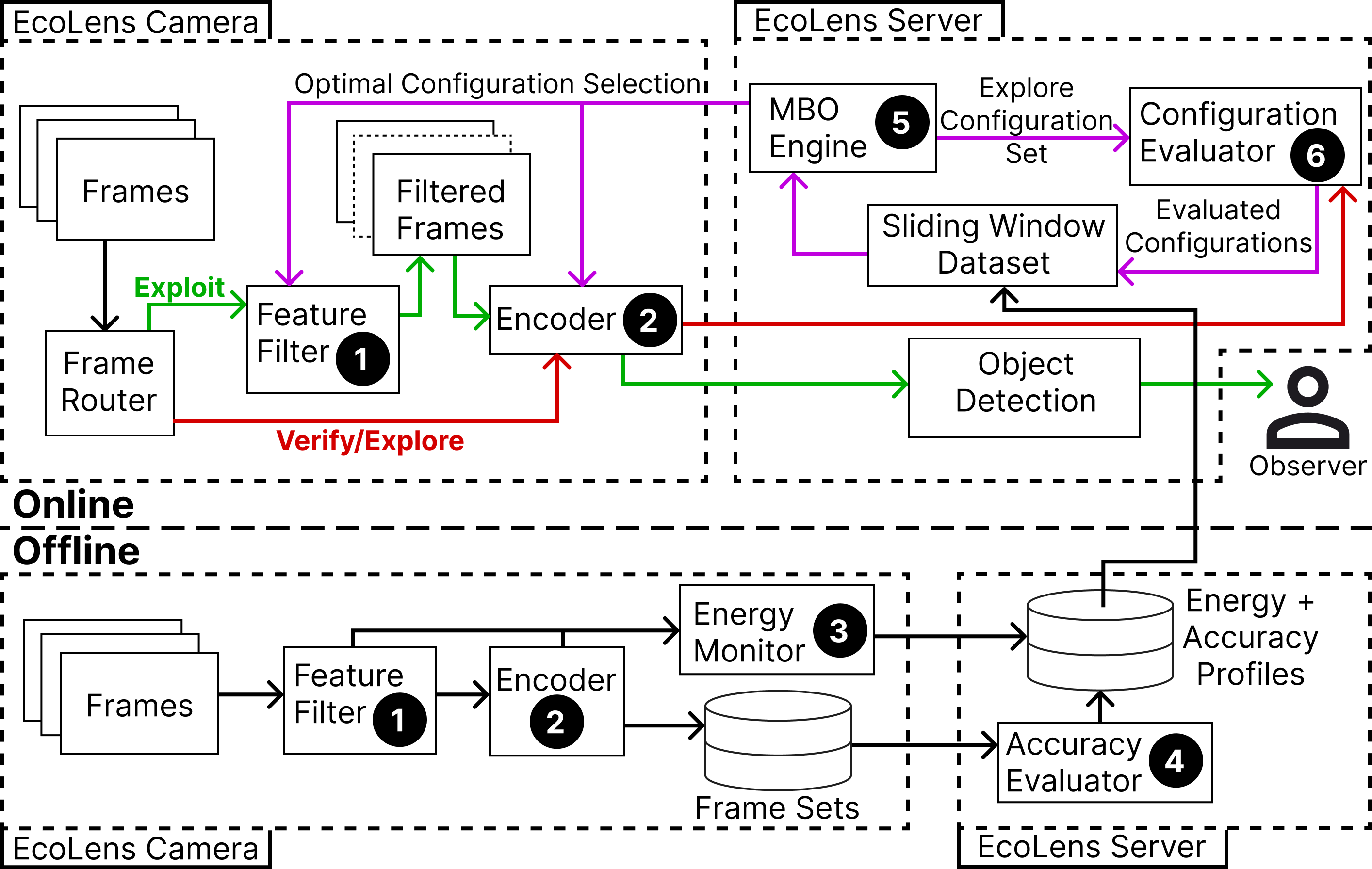}
    \caption{EcoLens System Overview}
    \label{fig:ecolens-design-diagram}
\end{figure}

Our proposed system solves the previously described challenges of 1) a complex impact of configurations on energy and accuracy, and 2) configuration performance drift after deployment. We solve this in a two stage process: an \textbf{offline sensing} stage where we gather a dataset to understand each configuration's impact on energy and accuracy, and an \textbf{online configuration adaptation} stage where we use this information to dynamically adjust the video processing configuration in a fast, online manner.
\newline\newline
\noindent\textit{Feature Filter \protect\designlabel{1}:}
The feature filter is responsible for selecting frames received from the camera based on the predefined filter threshold. First, the filter calculates the difference between frames based on the \texttt{pixel}, \texttt{area}, or \texttt{edge} features (described in \ref{filtering-background}). If the difference exceeds the filter threshold, the frame is passed to the encoder~\protect\designlabel{2} for further processing. During the online phase, the threshold is selected dynamically by the MBO engine~\protect\designlabel{5}.
\newline\newline
\textit{Encoder \protect\designlabel{2}:}
The encoder encodes raw frames using H.264. Once encoded, these frames can be transmitted to the server for further analysis. During the exploit phase, the encoding bitrate is selected by the MBO engine \protect\designlabel{5}. During the verify/explore phase, the frame bitrate is set to the maximum (3000 Kbps).
\newline\newline
\textit{Energy Monitor \protect\designlabel{3}:}
The energy monitor records the power consumption of the EcoLens camera during frame reading, encoding, and filtering. It stores the average energy consumption for each configuration in the energy profile.
\newline\newline
\textit{Accuracy Evaluator \protect\designlabel{4}:}
The accuracy evaluator takes all sets of frames generated by the offline configuration runs and calculates the object detection accuracy for each. It stores the average accuracy for each configuration in the accuracy profile. Initially, the online sliding window dataset is populated with Pareto-optimal configurations from the offline energy and accuracy profiles.
\newline\newline
\textit{MBO Engine \protect\designlabel{5}:}
The multi-objective Bayesian optimization (MBO) engine suggests the best set of configurations to profile to determine the optimal configuration. It uses a \textit{sliding window} approach to store observed configurations, maintaining a fixed-size dataset of the most recently evaluated configurations as input for the MBO model. This approach prioritizes recently observed data points, encouraging the model to generate diverse suggestions and adapt to changing conditions. In addition to MBO model suggestions, the engine augments its recommendations with high-performing configurations already present in the dataset -- specifically, those that lie on the Pareto front and meet the target accuracy threshold. The MBO model employs the Expected Hypervolume Improvement acquisition function~\cite{mboglobal} to expand the Pareto front across the objectives of accuracy and energy efficiency. Additionally, the engine selects an optimal configuration from the dataset for the exploit phase. This configuration meets the accuracy target, minimizes energy consumption, and has been evaluated in the most recent exploration round.
\newline\newline
\textit{Configuration Evaluator \protect\designlabel{6}:}
The configuration evaluator calculates the object detection accuracy of a given set of filtering/encoding configurations. The ground truth frames are received during the verify/explore phase of the online algorithm. Each configuration is evaluated by filtering and re-encoding the frames before performing model inference. Once the configurations have been evaluated, the results are stored in the sliding window dataset.

\subsection{Offline Sensing Stage}
\label{offlineprofiling}

\begin{figure*}[t]
    \centering
    \includegraphics[width=\textwidth]{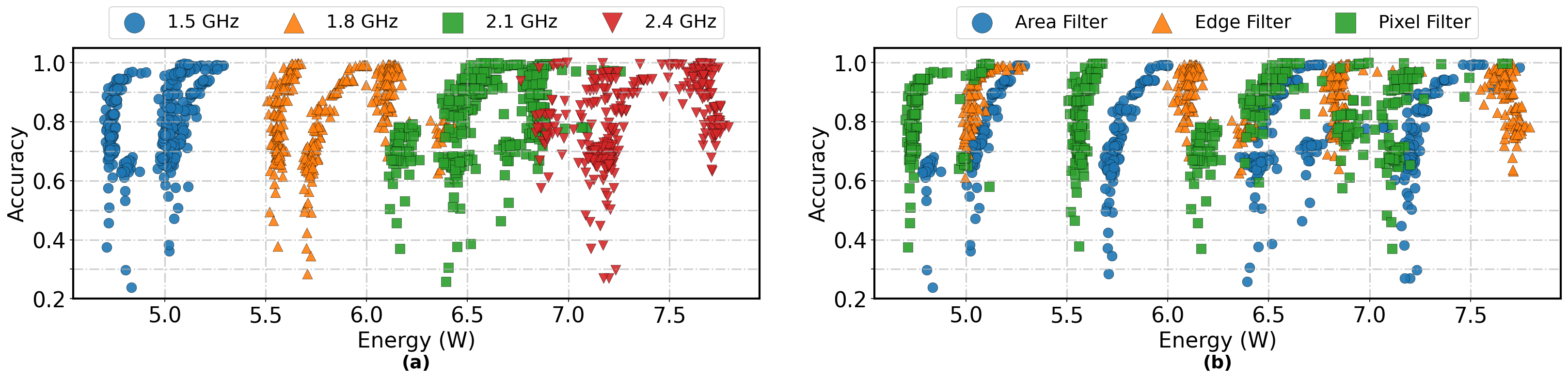}
    \caption{Plot of the offline energy and accuracy profiles from the Jackson Hole live camera feed during day time. Figure (a) highlights the impact of CPU frequency while figure (b) highlights the impact of frame filter.}
    \label{fig:energy-accuracy-JH-profile}
\end{figure*}

Through offline energy sensing and accuracy evaluation, we are able to gather key insights to address the complex impact of configurations on both energy and accuracy. The first is that in any scene, setting the device CPU frequency to 1.5 GHz (the minimum for the Raspberry Pi 5) and using the \texttt{pixel} feature filter yields the best energy/accuracy trade-off. This allows us to simplify the configuration search space. Second, we gather an energy profile of configurations that can be used as an initialization dataset for our adaptive online system. The remaining part of this section includes details of our offline sensing methodology.

We gathered a comprehensive offline profile of 1,452 configurations, shown in Table~\ref{tab:configurations}, on a varied set of video clips~\cite{jh_livecam, alma_livecam} using an energy sensing testbed. We captured one clip during the day, a highly dynamic scene with significant movement from cars and pedestrians, while the other was recorded at night, resulting in a more static scene with minimal object motion. The exact number of configurations can be adjusted for each camera deployment. A wider range can be evaluated for a larger online selection space, while a smaller range can be chosen to save deployment time.

Figure~\ref{fig:energy-accuracy-JH-profile} shows the profile of the daytime video. Figure \ref{fig:energy-accuracy-JH-profile}a plots energy consumption and accuracy for each tested streaming configuration. The data reveals distinct clusters corresponding to different frequency settings. It is evident that a CPU frequency of 1.5 GHz consistently consumes the least amount of energy across all configurations. Moreover, this frequency setting forms the entire Pareto front. In Figure \ref{fig:energy-accuracy-JH-profile}b, the same data is visualized to show the impact of different filters. Within the 1.5 GHz cluster, configurations using the \texttt{pixel} filter form the Pareto front, aligning with the preliminary results shown in Figure \ref{fig:motivation-1-single}. This reinforces our earlier observations that the \texttt{pixel} filter is the most energy-efficient choice. A similar pattern emerges when analyzing the nighttime video, despite its markedly different visual characteristics. The consistency of these trends suggests that our findings can be generalized to other camera feeds.

Previously the configuration search space consisted of four dimensional tuples, ($C$, $F$, $T$, $B$), where $C$ is CPU frequency, $F$ is feature filter, $T$ is filter threshold, and $B$ is video bitrate. Based on our findings, we greedily choose 1.5 GHz and the \texttt{pixel} filter for all streaming settings. This allows us to model the search for the optimal configuration as a two-dimensional MBO problem. We formally define the search space as $\mathbf{X}=\mathbf{T} \times \mathbf{B}$. $\mathbf{x} \in \mathbf{X}$ is a single configuration tuple containing the filter threshold and video bitrate. 

We define the objective functions as follows:
\begin{itemize}
    \item $A(\mathbf{x})$ is the IoU (Intersection over Union) of the video ground truth with the video using configuration $\mathbf{x}$.
    \item $E(\mathbf{x})$ is the average energy consumed on the edge device when processing the video using configuration $\mathbf{x}$.
\end{itemize}

Given a target accuracy, $A_\text{target}$, the optimization problem can be formalized as:

\vspace{-0.25em}
\begin{equation}
\begin{aligned}
    \min_{\mathbf{x}_{i,j} \in \mathbf{X}} \quad & \sum_{i=0}^{T} \sum_{j=100}^{B} E(\mathbf{x}_{i,j}) \\
    \text{s.t.} \quad & A(\mathbf{x}_{i,j}) \geq A_{\text{target}}
\end{aligned}
\end{equation}
\vspace{-1em}

\begin{table*}[t]
    \centering
    \caption{Evaluation of Various Processing Approaches Across Videos}
    \label{tab:results}
    \begin{tabular}{c c c c c c c}
        \toprule
        \textbf{Video} & \textbf{Method} & \textbf{Avg Energy (W)} & \textbf{Avg Accuracy (\%)} & \textbf{Norm. Energy} & \textbf{Energy Savings (\%)} \\
        \midrule \midrule
        \multirow{5}{*}{Jackson Hole Day~\cite{jh_livecam}} 
        & Reducto & 7.569 & 93.72 & 1.000 & 0.0 \\
        & Baseline & 7.223 & 100 & 0.954 & 4.6 \\
        & Offline & 4.808 & 91.69 & 0.635 & 36.5 \\
        & EcoLens & 4.731 & 89.22 & 0.625 & 37.5 \\
        \midrule
        \multirow{5}{*}{Jackson Hole Night~\cite{jh_livecam}} 
        & Reducto & 7.924 & 100 & 1.000 & 0.0 \\
        & Baseline & 7.039 & 100 & 0.888 & 11.2 \\
        & Offline & 5.046 & 92.48 & 0.637 & 36.3 \\
        & EcoLens & 4.582 & 87.31 & 0.578 & 42.2 \\
        \midrule
        \multirow{5}{*}{UIUC Campus~\cite{alma_livecam}} 
        & Reducto & 7.396 & 95.67 & 1.000 & 0.0 \\
        & Baseline & 7.245 & 100 & 0.980 & 2.0 \\
        & Offline & 4.115 & 87.98 & 0.556 & 44.4 \\
        & EcoLens & 4.086 & 88.85 & 0.552 & 44.6 \\
        \bottomrule
    \end{tabular}
\end{table*}

\begin{figure*}[t]
    \centering
    \includegraphics[width=\textwidth]{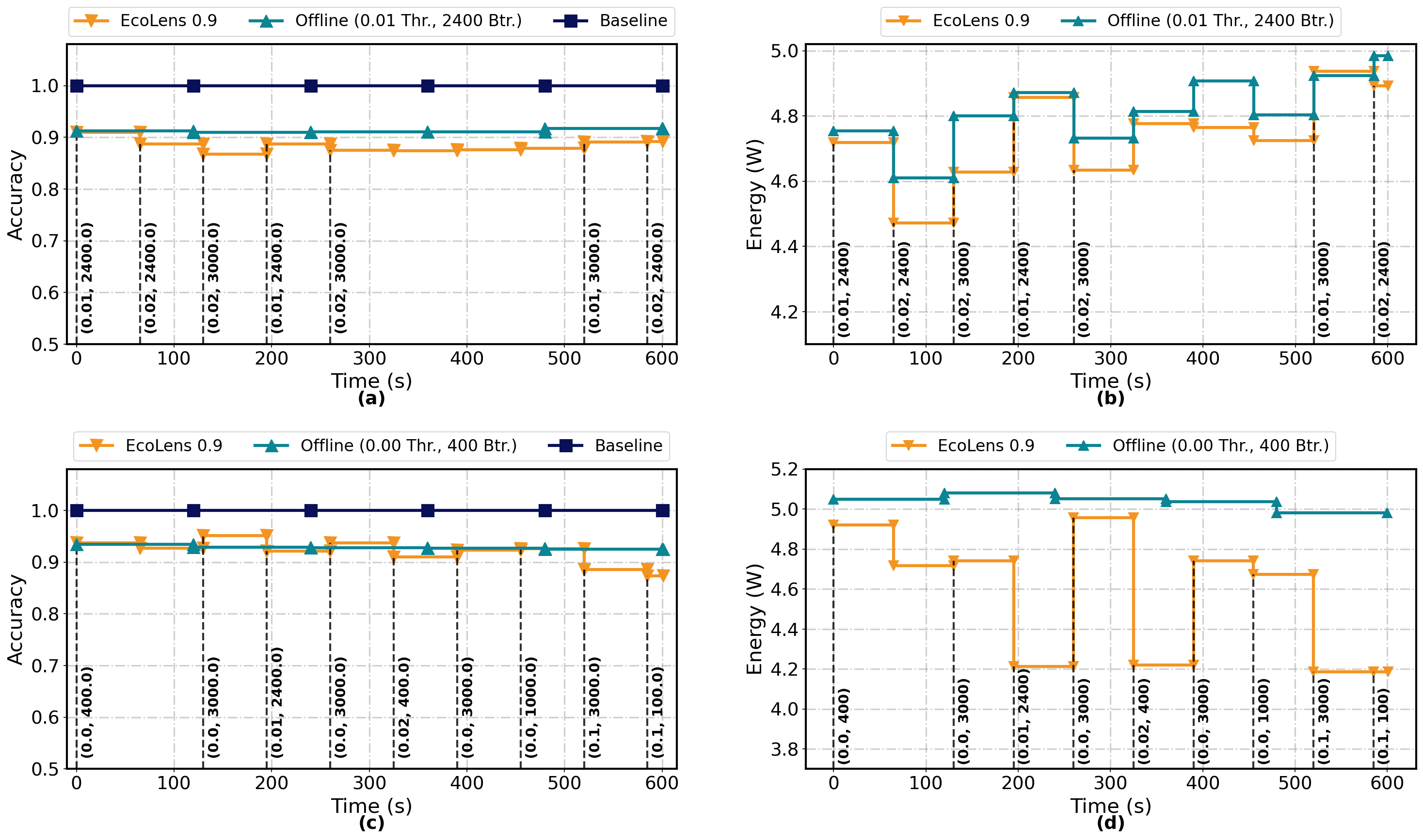}
    \caption{Accuracy and energy consumption of processing all frames without filtering, processing only based on the offline profiling optimal configuration, and processing using EcoLens with a 90\% accuracy target. Experiments (a) and (b) are conducted during the day while (c) and (d) are conducted during the night. The baseline energy is excluded from the graph to emphasize offline vs. EcoLens comparison, but is included in Table~\ref{tab:results}.}
    \label{fig:JH-eval-all}
\end{figure*}

\subsection{Online Adaptive Stage}

The next challenge to address is applying these insights in a system that quickly adapts the configuration to the video scene. As described in \ref{profiling-challenges}, performing an offline search of Table~\ref{tab:configurations} to find the best configuration is not optimal due to data drift. Our system needs to estimate the best configuration repeatedly, while streaming. We design an online algorithm that uses MBO to quickly find the best configuration based on recently observed data. This section describes the algorithm in detail.

The algorithm begins by sending a short burst of ground truth frames from the camera to the edge server at the maximum encoding bitrate without filtering. The configuration evaluator \protect\designlabel{6} receives this ground truth set and evaluates the currently selected optimal configuration in the \textbf{verify phase}. The newly calculated accuracy is updated in the sliding window dataset. This ensures that the optimal configuration is consistently reevaluated and updated accordingly if the observations change.

Next, the \textbf{explore phase} begins in the MBO engine~\protect\designlabel{5} by retrieving all datapoints currently within the sliding window.  From this set, a configurable number of Pareto optimal configurations are selected. If insufficient Pareto optimal points above the desired accuracy exist, the algorithm defaults to selecting the lowest energy points that still meet the minimum accuracy requirement. Next, a batch of suggestions is retrieved from the MBO model. These two sets of configurations are concatenated, and any duplicate entries are removed. The accuracies of the remaining configurations are then profiled on the edge server. The sliding window dataset is updated with the newly measured accuracy values and the most recently recorded energy consumption for each of these observations. Finally, the best configuration from these most recently profiled configurations is selected as the optimal configuration and sent to the camera.

Once the camera receives the newly chosen optimal configuration, the feature filter~\protect\designlabel{1} and encoder~\protect\designlabel{2} are updated to use this filter threshold and bitrate. The \textbf{exploit phase} begins, where the camera filters, encodes and sends the frames to the server as it would in a traditional object detection pipeline. During this phase, the camera takes advantage of the optimal energy and accuracy characteristics of the chosen configuration and imposes no additional overhead on the edge server. The server results can be queried by an observer. Once the exploit phase ends, the verify phase begins again.

\section{Implementation}

\subsection{Offline Sensing}
\label{offline-profiling-implementation}

We implement the sensing testbed in about 1k lines of \texttt{Python} and 500 lines of \texttt{C++}. The test bed was configured with a Raspberry Pi 5~\cite{rpi5} connected to a USB-C multimeter~\cite{tc66c} to measure energy consumption. The video processing pipeline was structured into four parallel threads: the first thread read saved video frames from disk and wrote them to shared memory for filtering; the second thread processed frames through the filtering mechanism, passing only those that met the criteria to encoding shared memory; the third thread encoded the filtered frames and stored them on the file system; and the fourth thread recorded energy measurements at 0.25-second intervals throughout the process. Each configuration generated a distinct set of frames, reflecting the impact of various filtering and encoding parameters.

We used an Nvidia 4060 RTX GPU running YOLOv8 models~\cite{yolo} to predict bounding boxes for each frame in the configuration set, and accuracy was assessed using the Intersection over Union (IoU) metric. Missing bounding boxes were assigned a score of zero. To facilitate temporal comparison, each frame was indexed based on its position in the original ground truth video. When a frame was absent due to filtering or dropping, the most recent previous frame was used for comparison against the corresponding ground truth frames. For example, if the configuration included frame 14 but omitted frames 15 and 16, frame 14 was compared against ground truth frames 14, 15, and 16 to estimate the accuracy impact of frame omission.

\subsection{Online System}
We implement the online system in about 1k lines of \texttt{Python} using the same hardware as the offline sensing testbed. The MBO engine is implemented using \texttt{Trieste-4.3.0} \cite{trieste}. The configuration evaluator uses the same temporal accuracy metric as described in \ref{offline-profiling-implementation}. We implement the EcoLens online stage using an explore time of 5 seconds and an exploit time of 60 seconds. We find that this setting reduces the overhead of configuration selection compared to the exploitation period and also allows enough re-profiling to adapt to changing video semantics. We choose a sliding window size of 20 observations and profile 10 configurations per explore phase (6 manual selections, 4 MBO selections).
\section{Evaluation and Results}

\subsection{Online Configuration Selection Time}

Our evaluation of EcoLens across all videos in our dataset resulted in an average overhead of 11.68 seconds for the MBO configuration selection. The XL (largest) model took an average of 32.92 seconds for configuration evaluation, while the Nano (smallest) model took an average of 19.56 seconds. In total, the average explore round took \textbf{44.60 seconds} with the XL model and \textbf{31.24 seconds} with the Nano model. This exploration time is short enough to support real-time operation, making online execution feasible. Additionally, the explore and exploit periods can be adjusted to meet individual system requirements. A longer exploration phase would introduce more overhead but provide a broader video context for selecting the optimal configuration. Due to the massive overhead of comprehensive profiling, we do not compare EcoLens to an oracle in our evaluation. Instead, we compare EcoLens to encoding all frames at the maximum bitrate (the accuracy `baseline'), our offline profiling without online configuration adaptation, and the filtering method used in Reducto\cite{reducto}.


\subsection{Accuracy Performance Analysis}

Figure \ref{fig:JH-eval-all} shows an in-depth evaluation of EcoLens with a target accuracy of 90\%. It compares EcoLens to purely offline profiling, i.e. choosing the best initial configuration based on the offline sensing described in section \ref{design}, and the baseline (2.4 GHz, \texttt{pixel} filter, 0.0 filter threshold, 3000 bitrate). The vertical lines show where EcoLens changes configurations, with the label describing the new threshold and bitrate. The evaluation shown in Figures \ref{fig:JH-eval-all}(a) and \ref{fig:JH-eval-all}(b) is conducted on the Jackson Hole live camera dataset during daytime, while Figures \ref{fig:JH-eval-all}(c) and \ref{fig:JH-eval-all}(d) are conducted on the Jackson Hole live camera during nighttime. The characteristics of each video are drastically different; during daytime, there are many dynamic objects like pedestrians and cars, while during the night, the scene is much more static.

As shown in both Figure \ref{fig:JH-eval-all}(a) and Figure \ref{fig:JH-eval-all}(c), EcoLens maintains accuracy at approximately 0.9, with minor fluctuations due to configuration updates. In both video scenes, the running accuracy does not drop below 5\% of the target. The offline approach maintains the target accuracy level of 0.9 for the duration of the video. While the offline approach maintains a more steady accuracy, it does not achieve as optimal an accuracy/energy trade-off as EcoLens. Additionally, the large difference in optimal configuration from offline sensing ([0.01 filter threshold, 2400 bitrate] compared to [0.00 filter threshold, 400 bitrate]) shows that the ideal configuration shifts over time. This further highlights the advantage of re-profiling, since as the scene changes EcoLens will continue to adapt to the most optimal configuration.

Table~\ref{tab:results} shows evaluation of Reducto~\cite{reducto}, baseline, offline, and EcoLens. EcoLens, offline, and Reducto were all given an accuracy target of 90\%. The evaluation shows that EcoLens does not drop more than 3\% below the target.

\subsection{Energy Performance Analysis}

Figures \ref{fig:JH-eval-all}(b) and \ref{fig:JH-eval-all}(d) reveal clear differences between the offline approach and EcoLens. EcoLens effectively reduces energy consumption by dynamically adjusting the bitrate and threshold in response to real-time conditions. In both \ref{fig:JH-eval-all}(b) and \ref{fig:JH-eval-all}(d), EcoLens demonstrates periodic energy reductions, with clear downward steps as lower-power configurations are selected. The most significant energy savings occur during the nighttime video, shown in \ref{fig:JH-eval-all}(d), where EcoLens consistently achieves energy levels below offline and baseline, even dropping below 4.2 W. This suggests that under different video conditions, EcoLens successfully identifies energy-efficient configurations without excessive accuracy loss.

Table~\ref{tab:results} shows a broader overview of energy savings. Across all videos, Reducto uses the most energy. This is because Reducto always uses the \texttt{edge} filter for object detection queries, which is the least energy-efficient choice. The baseline consistently uses the second most, followed by the offline method. EcoLens consistently outperforms all other methods in energy savings, trading off a small percentage of accuracy for a huge amount of energy savings. The energy savings of EcoLens range from 37.5\% -- 44.6\% compared to Reducto, and 34.5\% -- 43.6\% compared to the baseline.

\subsection{Discussion and Trade-offs}

Although our evaluation focuses on stationary cameras, our methodology readily generalizes to applications such as drones and wearable devices, potentially requiring a broader range of thresholds and bitrates during the sensing stage. The results demonstrate the trade-offs between energy efficiency and accuracy across all methods. The baseline configuration achieves the highest accuracy but at the cost of excessive energy consumption. Reducto optimizes for bandwidth and server load at the cost of edge device energy. The offline approach provides a stable balance between accuracy and energy usage but lacks real-time adaptability. By leveraging dynamic optimization, EcoLens achieves significant energy savings while maintaining accuracy within acceptable limits. This makes it particularly suitable for resource-constrained edge devices, where energy efficiency is critical. For these applications, EcoLens presents the most effective solution, as it successfully balances accuracy and power consumption based on real-time conditions.
\section{Conclusion}
In this paper, we identify that existing approaches do not optimize for energy efficiency in video streaming pipelines and gather a comprehensive profile of the impact of video processing configurations on edge device battery life. We design and propose an adaptive streaming system, EcoLens, that leverages offline sensing and MBO for online selection to reduce energy consumption while preserving inference accuracy. Finally, we conduct an extensive evaluation in multiple locations demonstrating the effectiveness of our approach in reducing edge device energy usage without significantly compromising analytical performance.

\bibliographystyle{IEEEtran}
\bibliography{references.bib}

\end{document}